\documentclass[letterpaper]{article}
\usepackage{spconf,amssymb,amsmath,graphicx,booktabs,adjustbox,enumitem}
\usepackage[table]{xcolor}
\usepackage[final]{microtype}
\usepackage{dblfloatfix}

\title{Cloth-HUGS: Cloth Aware Human Gaussian Splatting}
\name{Sadia Mubashshira, Nazanin Amini, Kevin Desai}
\address{
University of Texas at San Antonio, San Antonio, TX, USA}

\begin{document}
%
\maketitle
\begin{abstract}
We present Cloth-HUGS, a Gaussian Splatting based neural rendering framework for photorealistic clothed human reconstruction that explicitly disentangles body and clothing. Unlike prior methods that absorb clothing into a single body representation and struggle with loose garments and complex deformations, Cloth-HUGS represents the performer using separate Gaussian layers for body and cloth within a shared canonical space. The canonical volume jointly encodes body, cloth, and scene primitives and is deformed through SMPL-driven articulation with learned linear blend skinning weights. To improve cloth realism, we initialize cloth Gaussians from mesh topology and apply physics-inspired constraints, including simulation-consistency, ARAP regularization, and mask supervision. We further introduce a depth-aware multi-pass rendering strategy for robust body-cloth-scene compositing, enabling real-time rendering at over 60 FPS. Experiments on multiple benchmarks show that Cloth-HUGS improves perceptual quality and geometric fidelity over state-of-the-art baselines, reducing LPIPS by up to 28\% while producing temporally coherent cloth dynamics.
\end{abstract}  
\section{Introduction}
\label{Intro}

Photorealistic human avatars with controllable pose and clothing dynamics are fundamental to immersive applications such as virtual reality, telepresence, and digital content creation. Existing approaches for human avatar synthesis broadly fall into volumetric neural rendering, Gaussian-based avatar representations, and learning-based cloth modeling techniques with physics-inspired or simulation-based supervision.

Neural Radiance Field (NeRF)~\cite{mildenhall2020nerf} –based human rendering methods reconstruct high-fidelity performers from monocular videos by learning canonical volumetric representations animated via pose-conditioned deformation fields or skeletal transformations~\cite{weng2022humannerf, jiang2022neuman}. Despite a strong view-consistent appearance, these methods use costly volumetric ray marching and complex deformation mappings, resulting in slow training and rendering that limit scalability and real-time deployment.

To improve efficiency, recent work has shifted toward 3D Gaussian Splatting (3DGS)~\cite{kerbl20233d}, which replaces volumetric integration with explicit anisotropic Gaussian primitives rendered via rasterization. Several methods extend 3DGS to articulated humans using SMPL-driven deformation, enabling fast convergence and real-time rendering with competitive quality~\cite{kocabas2023hugs, hu2023gauhuman,lei2023gart}. However, these approaches typically model body and clothing as a single deformable layer, restricting geometric expressiveness for loose garments and preventing garment-level control such as cloth editing or virtual try-on.

In parallel, learning-based garment models trained with simulation-based or physics-inspired supervision explicitly capture cloth–body interactions and pose-dependent wrinkle formation~\cite{patel2020tailornet, santesteban2021self, santesteban2022snug}. Recent work such as PhysAvatar~\cite{zheng2024physavatar} integrates physical simulation into layered human reconstruction, but relies on mesh-based pipelines and costly optimization, making it difficult to integrate into efficient neural rendering frameworks.

More recent layered avatar representations attempt to disentangle clothing from the body using Gaussian primitives or explicit meshes~\cite{clocapgs, wang_cloth, reloo}, highlighting the importance of factorized modeling. Nevertheless, these methods often lack explicit physics-inspired supervision for cloth deformation and robust depth-consistent compositing between body, cloth, and scene layers, leading to instability under complex occlusions.

To address these challenges, we propose \textit{Cloth-HUGS}, a cloth-aware Gaussian Splatting framework that explicitly disentangles body and cloth into separate canonical Gaussian layers while retaining the efficiency of 3DGS. Cloth Gaussians are initialized and supervised using pseudo-ground-truth cloth
geometry generated by SNUG~\cite{santesteban2022snug}, a pretrained garment model. We learn gaussian parameters with physics-inspired supervision, including simulation alignment, as-rigid-as-possible (ARAP) regularization, and cloth-specific linear blend skinning (LBS) constraints. We further introduce a depth-aware multi-pass rendering strategy for robust body–cloth–scene compositing. Cloth-HUGS enables photorealistic clothed human synthesis with controllable garment dynamics and efficient training ($\sim$40 minutes).

\begin{figure*}[t]
    \centering
    \includegraphics[width=.95\textwidth]{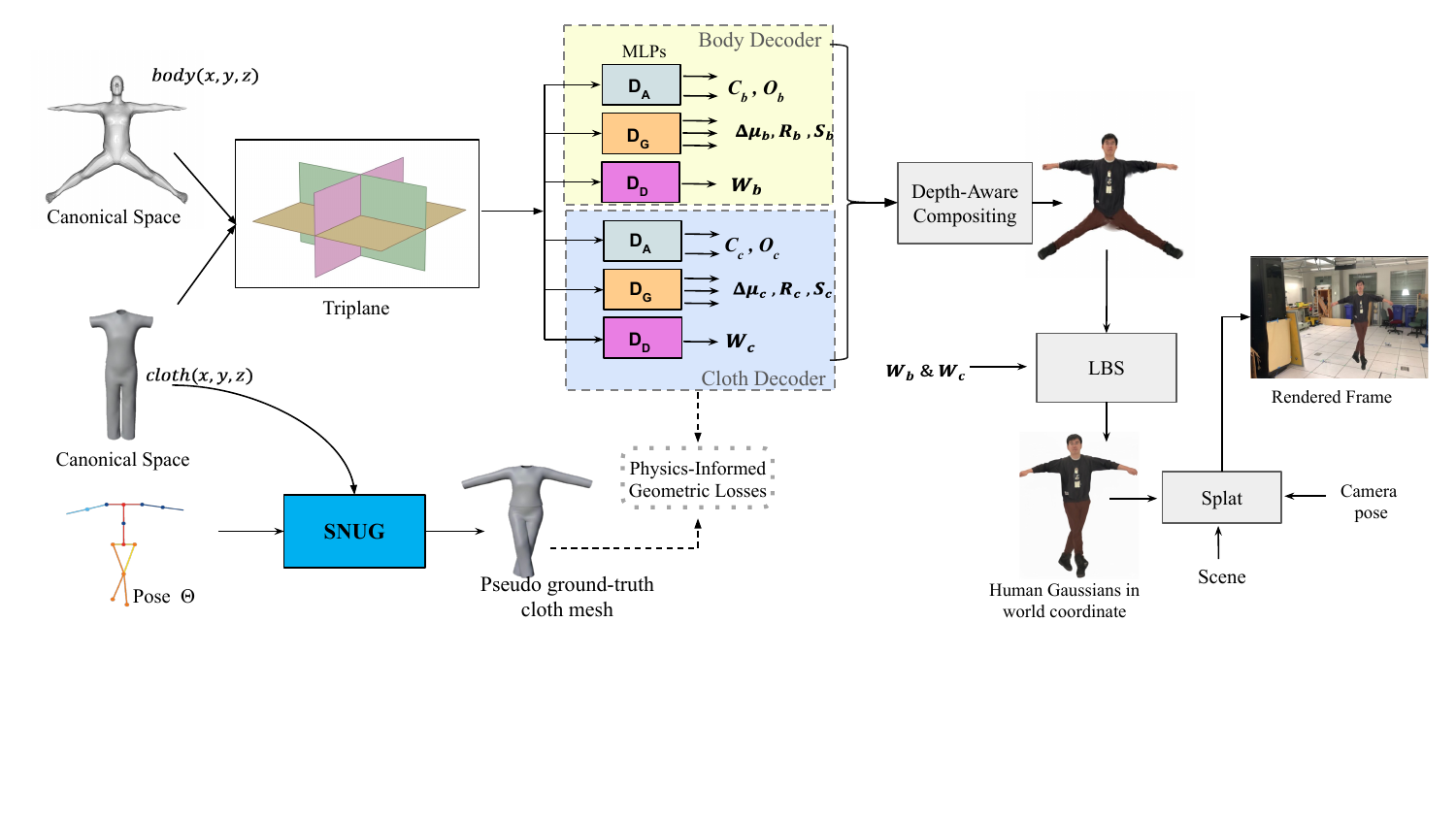}
    \vspace{-26mm}
    \caption{\textbf{Cloth-HUGS Overview}. From a monocular video, we build a body–cloth disentangled Gaussian avatar using SMPL-conditioned TriPlane features, physics-inspired cloth refinement, and depth-aware rendering.
}
    \vspace{-4mm}
    \label{fig:architecture}
\end{figure*}



Our main contributions are as follows:
\begin{itemize}[topsep=2pt, itemsep=2pt, parsep=0pt]
    \item We introduce Cloth-HUGS, a cloth-aware Gaussian Splatting framework that explicitly disentangles body and cloth for high-fidelity garment deformation.
    \item We integrate physics-guided supervision, including simulation alignment, ARAP, and cloth-specific LBS regularization, improving temporal stability and realism.
    \item We propose a depth-aware multi-pass rendering pipeline to handle body–cloth–scene occlusions, enabling applications such as cloth editing and virtual try-on.
    \item Cloth-HUGS achieves state-of-the-art performance across diverse motions, reducing LPIPS by 28\% on average on NeuMan~\cite{jiang2022neuman} while consistently improving PSNR, SSIM, and FID.
\end{itemize}

\vspace{-3mm}
\section{Method}
\label{sec:method}
\vspace{-1mm}

Given a monocular RGB video $\{I_t\}_{t=1}^T$ of a moving human in a static scene, along with per-frame camera parameters $\{K_t, R_t, T_t\}$ and estimated SMPL pose parameters $\{\beta, \theta_t\}$, our goal is photorealistic novel-view synthesis and pose-controllable animation of clothed humans.

Cloth-HUGS builds on 3D Gaussian Splatting (3DGS)~\cite{kerbl20233d}, which represents scenes using anisotropic 3D Gaussian primitives parameterized by center $\mu$, opacity $o$, rotation $R$, scale $S$, and spherical-harmonic coefficients for view-dependent appearance. Gaussians are projected to the image plane and rendered via differentiable alpha compositing, enabling efficient training and real-time rendering. We adopt the SMPL parametric body model~\cite{loper2015smpl} to represent human shape and pose using learned blend shapes and Linear Blend Skinning (LBS), providing a compact, controllable body prior but unable to model cloth dynamics.

Cloth-HUGS represents body, cloth, and scene as separate Gaussian layers in a shared canonical space (Fig.~1), with the cloth layer modeled as an explicit geometric representation articulated via SMPL using learned LBS weights. To guide cloth geometry, we leverage per-frame cloth meshes $\{M_t^c\}$ generated by SNUG~\cite{santesteban2022snug} as auxiliary supervision. A depth-aware multi-pass rendering strategy enforces correct body–cloth–scene compositing and improves temporal consistency under occlusions.

We next describe our cloth-aware canonical representation and deformation (Sec.~\ref{sec:canonical_deformation}), cloth-based supervision (Sec.~\ref{sec:physics_supervision}), and depth-aware rendering pipeline (Sec.~\ref{sec:rendering}).

\vspace{-3mm}
\subsection{Cloth-Aware Representation and Deformation}
\label{sec:canonical_deformation}
\vspace{-1mm}

We disentangle the clothed human into \textit{body} and \textit{cloth} Gaussian sets in a shared canonical space, enabling cloth-specific motion, deformation, and depth layering while retaining SMPL-driven animatability. This contrasts with prior monolithic body-cloth representations~\cite{kocabas2023hugs,hu2023gauhuman,lei2023gart} and supports physics-guided cloth modeling.

\noindent\textbf{Initialization.} The body Gaussians $\mathcal{G}^{\text{canon}}_B=\{G^B_i\}_{i=1}^{N_B}$ are initialized from SMPL~\cite{loper2015smpl} in the canonical pose, with centers $\boldsymbol{\mu}^B_i$, rotations $\mathbf{R}^B_i$, and anisotropic scales $\mathbf{s}^B_i$ derived from vertex positions, surface normals, and local edge lengths. The cloth Gaussians $\mathcal{G}^{\text{canon}}_C=\{G^C_i\}_{i=1}^{N_C}$ are initialized analogously from SNUG-generated cloth meshes $\mathcal{M}^c$~\cite{santesteban2022snug} in a T-pose. For both sets, opacity and spherical-harmonic coefficients are initialized to neutral values.

We encode body and cloth properties using an explicit–implicit formulation based on triplane features. Body and cloth are assigned separate triplane encodings in a canonical space, with the cloth triplane modeling upper and lower garments for coherent dynamics. Each triplane consists of three feature planes $\mathbf{F}_{XY},\mathbf{F}_{XZ},\mathbf{F}_{YZ}\in\mathbb{R}^{256\times256\times32}$, from which features are bilinearly sampled and concatenated into $\mathbf{f}_{\text{tri}}(\mathbf{p})\in\mathbb{R}^{96}$ at a query point $\mathbf{p}$.

\noindent\textbf{TriPlane Feature Decoding and Deformation.}
Given triplane features $\mathbf{f}_{\text{tri}}(\mathbf{p})$, three MLP decoders predict Gaussian appearance, geometry, and deformation attributes.  
The appearance decoder $D_A$ outputs spherical-harmonic coefficients $\mathbf{c}_{\text{SH}}\in\mathbb{R}^{16\times3}$ and opacity $o=\sigma(\hat{o})$.  
The geometry decoder $D_G$ predicts position, rotation, and scale corrections $(\Delta\boldsymbol{\mu}, \Delta\mathbf{r}, \Delta\mathbf{s})$, yielding deformed Gaussian parameters

\begin{equation*}
\small
\begin{aligned}
\boldsymbol{\mu}_{\text{def}} &= \boldsymbol{\mu}_{\text{canon}} + \Delta\boldsymbol{\mu}, \\
\mathbf{R}_{\text{def}} &= \mathbf{R}_{\text{canon}}\mathcal{R}(\Delta\mathbf{r}), \\
\mathbf{s}_{\text{def}} &= \mathbf{s}_{\text{canon}} \odot \exp(\Delta\mathbf{s})
\end{aligned}
\end{equation*}

where $\mathcal{R}(\cdot)$ converts the 6D rotation representation to $SO(3)$ via Gram–Schmidt~\cite{zhou2019continuity}.  
The deformation decoder $D_D$ predicts Linear Blend Skinning weights $\mathbf{w}\in\mathbb{R}^{24}$ (softmax-normalized) and pose-dependent offsets $\Delta\mathbf{p}\in\mathbb{R}^{207\times3}$, modeling non-rigid effects such as muscle deformation and cloth wrinkles~\cite{loper2015smpl}. The predicted weights articulate $\boldsymbol{\mu}_{\text{def}}$ to world space via the SMPL skeleton.

\vspace{-3mm}
\subsection{Cloth-based Supervision}
\label{sec:physics_supervision}
\vspace{-1mm}


\noindent\textbf{Cloth LBS Regularization.}
We regularize predicted cloth Linear Blend Skinning (LBS) weights using reference weights transferred from the SMPL body, following prior skinning-weight transfer methods~\cite{patel2020tailornet,santesteban2022snug}. Ground-truth weights $\mathbf{W}_{\text{cloth}}^{\text{gt}}$ are obtained by mapping cloth vertices in T-pose to nearest SMPL vertices. The loss is
\begin{equation}
\mathcal{L}_{\text{cloth-lbs}} =
\lambda_{\text{cloth-lbs}}
\|\mathbf{W}_{\text{cloth}} - \mathbf{W}_{\text{cloth}}^{\text{gt}}\|_2^2.
\end{equation}

\noindent\textbf{Physics-Informed Geometric Losses.}
We further impose three complementary geometric constraints:

\begin{enumerate}
\item \textbf{Simulation Alignment ($\mathcal{L}_{\text{sim}}$).}
We align predicted cloth geometry with SNUG-generated meshes~\cite{santesteban2022snug} using a bidirectional Chamfer distance:
\begin{multline}
\mathcal{L}_{\text{sim}}
= \frac{\lambda_{\text{sim}}}{2} \Bigg[
\frac{1}{|\mathcal{V}_{\text{pred}}|}
\sum_{\mathbf{v}_i \in \mathcal{V}_{\text{pred}}}
\rho\!\left(
\min_{\mathbf{u}_j \in \mathcal{V}_{\text{gt}}}
\|\mathbf{v}_i - \mathbf{u}_j\|_2
\right) \\
+ \frac{1}{|\mathcal{V}_{\text{gt}}|}
\sum_{\mathbf{u}_j \in \mathcal{V}_{\text{gt}}}
\rho\!\left(
\min_{\mathbf{v}_i \in \mathcal{V}_{\text{pred}}}
\|\mathbf{u}_j - \mathbf{v}_i\|_2
\right)
\Bigg],
\end{multline}
where $\rho(\cdot)$ is the Geman-McClure function.

\item \textbf{ARAP Regularization ($\mathcal{L}_{\text{ARAP}}$).}
To preserve local cloth structure, we apply an As-Rigid-As-Possible constraint:
\begin{equation}
\mathcal{L}_{\text{ARAP}} =
\lambda_{\text{ARAP}}\,
\text{Var}\!\Big(
\{\|\mathbf{v}_i - \mathbf{v}_j\|_2 : (i,j)\in\mathcal{E}\}
\Big).
\end{equation}

\item \textbf{Mask Consistency ($\mathcal{L}_{\text{mask}}$).}
We enforce silhouette consistency between rendered and ground-truth cloth masks:
\begin{equation}
\mathcal{L}_{\text{mask}} =
\lambda_{\text{mask}}\,
\frac{1}{|N|}
\|\mathbf{M}_{\text{render}} - \mathbf{M}_{\text{gt}}\|_2^2.
\end{equation}
\end{enumerate}

\noindent\textbf{Combined Loss:}
\begin{equation}
\begin{split}
\mathcal{L} =&\;
\mathcal{L}_{\text{rec}} +
\lambda_{\text{cloth-lbs}}\mathcal{L}_{\text{cloth-lbs}} +
\lambda_{\text{sim}}\mathcal{L}_{\text{sim}} \\
&+
\lambda_{\text{ARAP}}\mathcal{L}_{\text{ARAP}} +
\lambda_{\text{mask}}\mathcal{L}_{\text{mask}},
\end{split}
\label{eq:total_loss}
\end{equation}
\vspace{-1mm}
with
\vspace{-3mm}
\begin{equation}
\mathcal{L}_{\text{rec}} =
\lambda_{\text{L1}}\mathcal{L}_{\text{L1}} +
\lambda_{\text{SSIM}}\mathcal{L}_{\text{SSIM}} +
\lambda_{\text{LPIPS}}\mathcal{L}_{\text{LPIPS}}.
\label{eq:reconstruction_loss}
\end{equation}

\vspace{-2mm}
\subsection{Depth-Aware Multi-Pass Rendering}
\label{sec:rendering}
\vspace{-1mm}

To handle occlusions between body, cloth, and scene, we employ a \textit{depth-aware multi-pass rendering} pipeline built on differentiable 3D Gaussian rasterization~\cite{kerbl20233d}. In the first pass, body and scene Gaussians are jointly rendered using standard alpha blending to produce a base image $\mathbf{I}_{\text{base}}$, where depth testing ensures correct inter-layer occlusion.
In the second pass, cloth Gaussians are rendered to obtain $\mathbf{I}_{\text{cloth}}$ using spherical-harmonic (SH) coefficients. Cloth visibility is determined via a depth matte $\mathbf{V}_{\text{cloth}}\in[0,1]^{H\times W}$, computed by jointly rendering cloth and scene Gaussians with binary color assignment and depth-aware compositing.

The final image is obtained as
\vspace{-2mm}
\begin{equation}
\mathbf{I}_{\text{final}} =
\mathbf{I}_{\text{cloth}} \odot \mathbf{V}_{\text{cloth}} +
\mathbf{I}_{\text{base}} \odot (1 - \mathbf{V}_{\text{cloth}}),
\vspace{-2mm}
\end{equation}

where $\odot$ denotes element-wise multiplication. This ensures correct cloth occlusion while allowing semi-transparent blending. Scene-only views are rendered independently preserving the efficiency of Gaussian Splatting while enabling physically consistent body-cloth-scene compositing.

\vspace{-3mm}
\section{Experimental Setup}
\label{sec:experiments}
\vspace{-2mm}
This section outlines our evaluation setup for Cloth-HUGS, including implementation details and training configuration (\ref{sec:impl-details}), benchmark datasets (\ref{sec:datasets}), and evaluation metrics (\ref{sec:metrics}).

\begin{figure*}[t]
    \includegraphics[width=.8\textwidth]{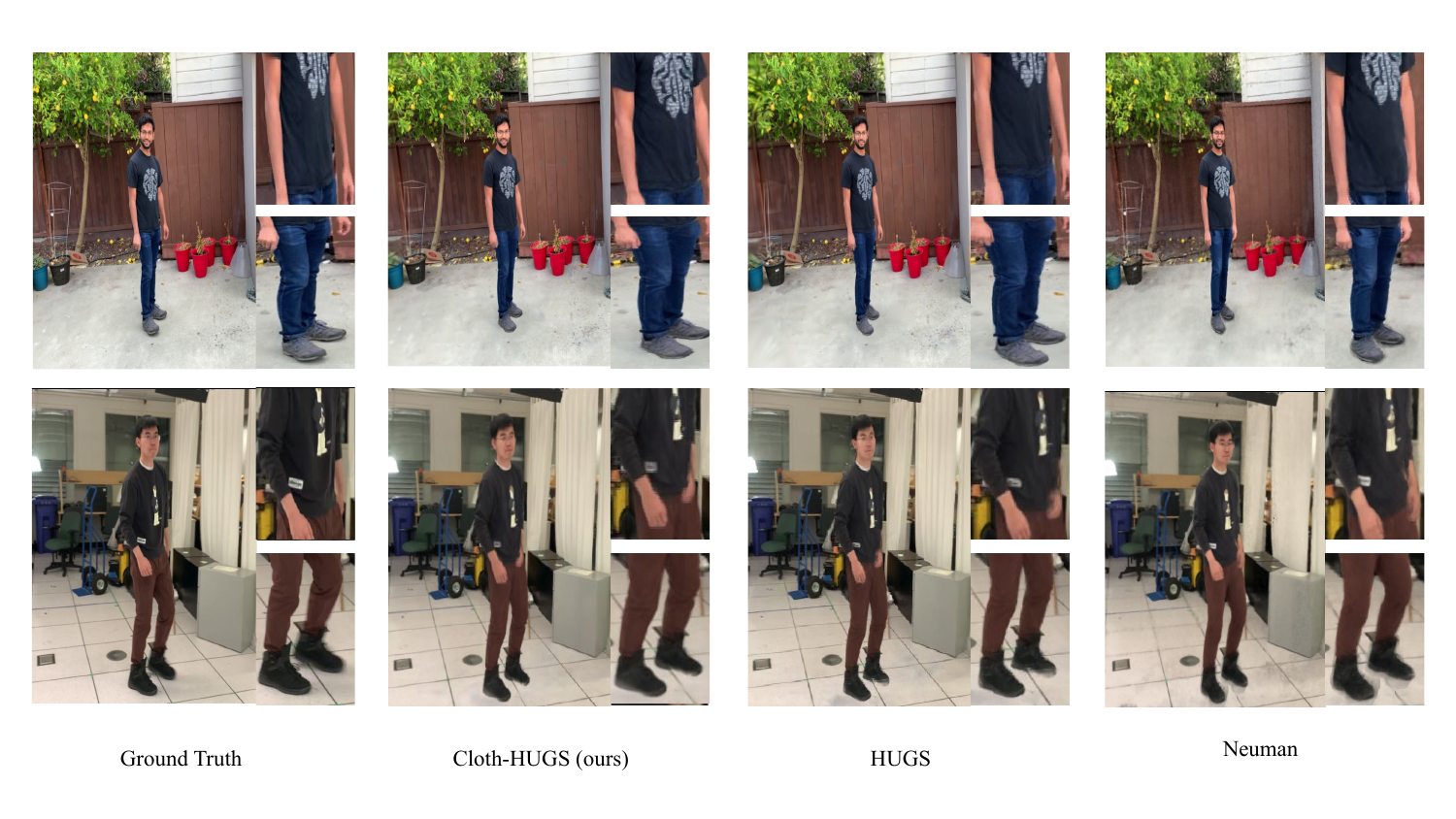}
    \centering
    \vspace{-5mm}
    \caption{
        Qualitative comparison with HUGS~\cite{kocabas2023hugs} and NeuMan~\cite{jiang2022neuman} with zoomed-in crops.
        Our Cloth-HUGS framework produces sharper details and more realistic cloth deformation, 
        accurately capturing wrinkles, folds, and body–cloth occlusions.
    }
    \label{fig:qualitative}
\vspace{-3mm}
\end{figure*}

\vspace{-2mm}
\subsection{Implementation Details}
\label{sec:impl-details}
\vspace{-1mm}

\noindent\textbf{Loss weights.}
We set the weights in Eqs.~\ref{eq:total_loss} and \ref{eq:reconstruction_loss} to
\(\lambda_{\text{L1}}{=}0.8\),
\(\lambda_{\text{SSIM}}{=}0.2\),
\(\lambda_{\text{LPIPS}}{=}1.0\),
\(\lambda_{\text{sim}}{=}1.0\),
\(\lambda_{\text{ARAP}}{=}0.5\),
\(\lambda_{\text{mask}}{=}1.0\),
and \(\lambda_{\text{cloth-lbs}}{=}1000.0\).

\noindent\textbf{Learning rates.}
Gaussian positions start at \(1.6\times10^{-4}\) and decay to \(1.6\times10^{-6}\) over 20k iterations.
Rotation, scale and opacity use fixed learning rates of
\(1.0\times10^{-3}\), \(5.0\times10^{-3}\), and \(5.0\times10^{-2}\), respectively.


\noindent\textbf{Runtime.}
Models are trained for 20k iterations at \(512{\times}512\) on an NVIDIA L40S GPU.
Sequences converge in $\sim$40 minutes, comparable to HUGS~\cite{kocabas2023hugs} with added cloth supervision.

\vspace{-2mm}
\subsection{Datasets}
\label{sec:datasets}
\vspace{-1mm}

\noindent\textbf{NeuMan}~\cite{jiang2022neuman}
contains six monocular handheld sequences (\textit{Seattle}, \textit{Citron},
\textit{Parking}, \textit{Bike}, \textit{Jogging}, \textit{Lab}) of 10-20 seconds, supporting multi-view reconstruction. We adopt 80\% of the frames for training, 10\% for validation, and 10\% for
testing.

\noindent\textbf{ZJU-MoCap}~\cite{peng2021neural}
provides calibrated multi-view captures in a controlled studio setting.
Following prior work~\cite{hu2023gauhuman,lei2023gart}, we evaluate subjects 377, 386, 387, 392, 393, and 394.
We use “camera 1” as the input view and hold out the remaining 22 cameras for evaluation.
We use the provided poses, masks, and camera calibration for training and testing.

\vspace{-2mm}
\subsection{Metrics}
\label{sec:metrics}
\vspace{-1mm}
We report PSNR, SSIM, and LPIPS to measure pixel fidelity, structural similarity, and perceptual quality, respectively.
We additionally report FID~\cite{heusel2017gans} to quantify distributional differences between rendered and real images.


\begin{table*}[b]
\centering

\resizebox{\textwidth}{!}{%
\begin{tabular}{l|ccc|ccc|ccc|ccc|ccc|ccc||ccc}
\toprule
 & \multicolumn{3}{c|}{Seattle} & \multicolumn{3}{c|}{Citron} & \multicolumn{3}{c|}{Parking} & \multicolumn{3}{c|}{Bike} & \multicolumn{3}{c|}{Jogging} & \multicolumn{3}{c||}{Lab} & \multicolumn{3}{c}{Average} \\
\cmidrule(lr){2-4}\cmidrule(lr){5-7}\cmidrule(lr){8-10}\cmidrule(lr){11-13}\cmidrule(lr){14-16}\cmidrule(lr){17-19}\cmidrule(lr){20-22}
Method & PSNR & SSIM & LPIPS & PSNR & SSIM & LPIPS & PSNR & SSIM & LPIPS & PSNR & SSIM & LPIPS & PSNR & SSIM & LPIPS & PSNR & SSIM & LPIPS & PSNR & SSIM & LPIPS \\
\midrule
NeRF-T & 21.84 & 0.69 & 0.37 & 12.33 & 0.49 & 0.65 & 21.98 & 0.69 & 0.46 & 21.16 & 0.71 & 0.36 & 20.63 & 0.53 & 0.49 & 20.52 & 0.75 & 0.39 &19.743 &0.643 &0.453\\

HyperNeRF & 16.43 & 0.43 & 0.40 & 16.81 & 0.41 & 0.56 & 16.04 & 0.38 & 0.62 & 17.64 & 0.42 & 0.43 & 18.52 & 0.39 & 0.52 & 16.75 & 0.51 & 0.23 &17.032 &0.423 &0.460 \\

Vid2Avatar & 17.41 & \cellcolor{yellow!25}0.56 & 0.60 & 14.32 & \cellcolor{yellow!25}0.62 & 0.65 & 21.56 & 0.69 & 0.50 & 14.86 & 0.51 & 0.69 & 15.04 & 0.41 & 0.70 & 13.96 & \cellcolor{yellow!25}0.60 & 0.68 & 16.192 & 0.565 & 0.637\\

NeuMan & \cellcolor{yellow!25}23.99 & \cellcolor{orange!25}0.78 & \cellcolor{yellow!25}0.26 & \cellcolor{yellow!25}24.63 & \cellcolor{orange!25}0.81 & \cellcolor{yellow!25}0.26 & \cellcolor{yellow!25}25.43 & \cellcolor{yellow!25}0.80 & \cellcolor{yellow!25}0.31 & \cellcolor{orange!25}25.55 & \cellcolor{yellow!25}0.83 & \cellcolor{yellow!25}0.23 & \cellcolor{yellow!25}22.70 & \cellcolor{yellow!25}0.68 & \cellcolor{yellow!25}0.32 & \cellcolor{yellow!25}24.96 & \cellcolor{orange!25}0.86 & \cellcolor{yellow!25}0.21 
&\cellcolor{yellow!25}24.543 &\cellcolor{yellow!25}0.793 &\cellcolor{yellow!25}0.265\\

HUGS & \cellcolor{orange!25}25.94 & \cellcolor{red!25}0.85 & \cellcolor{orange!25}0.13 & \cellcolor{orange!25}25.54 & \cellcolor{red!25}0.86 & \cellcolor{orange!25}0.15 & \cellcolor{red!25}26.86 & \cellcolor{red!25}0.85 & \cellcolor{orange!25}0.22 & \cellcolor{yellow!25}25.46 & \cellcolor{orange!25}0.84 & \cellcolor{orange!25}0.13 & \cellcolor{red!25}23.75 & \cellcolor{red!25}0.78 & \cellcolor{orange!25}0.22 & \cellcolor{orange!25}26.00 & \cellcolor{red!25}0.92 & \cellcolor{orange!25}0.09 &\cellcolor{orange!25}25.592 &\cellcolor{red!25}0.850 &\cellcolor{orange!25}0.157 \\
\rowcolor{gray!5}
\midrule

Ours & \cellcolor{red!25}26.15 & \cellcolor{red!25}0.85 & \cellcolor{red!25}0.10 & \cellcolor{red!25}25.78 & \cellcolor{red!25}0.86 & \cellcolor{red!25}0.09 & \cellcolor{orange!25}26.78 & \cellcolor{orange!25}0.84 & \cellcolor{red!25}0.14 & \cellcolor{red!25}25.56 & \cellcolor{red!25}0.85 & \cellcolor{red!25}0.10 & \cellcolor{orange!25}23.57 & \cellcolor{orange!25}0.76 & \cellcolor{red!25}0.18 & \cellcolor{red!25}26.14 & \cellcolor{red!25}0.92 & \cellcolor{red!25}0.07 &\cellcolor{red!25}25.663 &\cellcolor{orange!25}0.847 &\cellcolor{red!25}0.113 \\
\bottomrule
\end{tabular}
}

\vspace{-3mm}
\caption{Comparison of Cloth-HUGS with prior work on NeuMan \cite{jiang2022neuman} test images for \textit{both human and the scene regions} using PSNR↑, SSIM↑, and LPIPS↓. Higher is better for PSNR and SSIM. Lower is better for LPIPS.}
\label{tab:neuman_full}
\vspace{-2mm}
\end{table*}



\begin{table*}[b]
\centering
\resizebox{\textwidth}{!}{%
\begin{tabular}{l|ccc|ccc|ccc|ccc|ccc|ccc||ccc}
\toprule
 & \multicolumn{3}{c|}{Seattle} 
 & \multicolumn{3}{c|}{Citron} 
 & \multicolumn{3}{c|}{Parking} 
 & \multicolumn{3}{c|}{Bike} 
 & \multicolumn{3}{c|}{Jogging} 
 & \multicolumn{3}{c||}{Lab}
 & \multicolumn{3}{c}{Average} \\
\cmidrule(lr){2-4}\cmidrule(lr){5-7}\cmidrule(lr){8-10}\cmidrule(lr){11-13}\cmidrule(lr){14-16}\cmidrule(lr){17-19}\cmidrule(lr){20-22}
Method 
& PSNR & SSIM & LPIPS
& PSNR & SSIM & LPIPS
& PSNR & SSIM & LPIPS
& PSNR & SSIM & LPIPS
& PSNR & SSIM & LPIPS
& PSNR & SSIM & LPIPS
& PSNR & SSIM & LPIPS \\
\midrule

Vid2Avatar 
& 16.90 & 0.51 & 0.27 
& 15.96 & 0.59 & 0.28 
& 18.51 & 0.65 & 0.26 
& \cellcolor{yellow!25}12.44 & 0.39 & 0.54 
& 16.36 & 0.46 & 0.30 
& 15.99 & \cellcolor{yellow!25}0.62 & 0.34 & 16.027 & 0.537 & 0.331\\

NeuMan 
& \cellcolor{yellow!25}18.42 & \cellcolor{yellow!25}0.58 & \cellcolor{yellow!25}0.20 
& \cellcolor{yellow!25}18.39 & \cellcolor{yellow!25}0.64 & \cellcolor{yellow!25}0.19 
& \cellcolor{yellow!25}17.66 & \cellcolor{yellow!25}0.66 & \cellcolor{yellow!25}0.24 
& \cellcolor{orange!25}19.05 & \cellcolor{yellow!25}0.66 & \cellcolor{yellow!25}0.21 
& \cellcolor{red!25}17.57 & \cellcolor{yellow!25}0.54 & \cellcolor{yellow!25}0.29 
& \cellcolor{yellow!25}18.76 & \cellcolor{orange!25}0.73 & \cellcolor{yellow!25}0.23 & \cellcolor{yellow!25}18.308 & \cellcolor{yellow!25}0.635 & \cellcolor{yellow!25}0.227 \\

HUGS 
& \cellcolor{red!25}19.06 & \cellcolor{red!25}0.67 & \cellcolor{orange!25}0.15 
& \cellcolor{red!25}19.16 & \cellcolor{red!25}0.71 & \cellcolor{orange!25}0.16 
& \cellcolor{red!25}19.44 & \cellcolor{red!25}0.73 & \cellcolor{orange!25}0.17 
& \cellcolor{red!25}19.48 & \cellcolor{red!25}0.67 & \cellcolor{orange!25}0.18 
& \cellcolor{orange!25}17.45 & \cellcolor{red!25}0.59 & \cellcolor{orange!25}0.27 
& \cellcolor{orange!25}18.79 & \cellcolor{red!25}0.76 & \cellcolor{orange!25}0.18 & \cellcolor{red!25}18.897 & \cellcolor{red!25}0.688 & \cellcolor{orange!25}0.185 \\

\rowcolor{gray!5}
\midrule
Ours 
& \cellcolor{orange!25}18.68 & \cellcolor{orange!25}0.65 & \cellcolor{red!25}0.14 
& \cellcolor{orange!25}19.12 & \cellcolor{orange!25}0.70 & \cellcolor{red!25}0.13 
& \cellcolor{orange!25}19.32 & \cellcolor{orange!25}0.71 & \cellcolor{red!25}0.15 
& \cellcolor{red!25}19.48 & \cellcolor{orange!25}0.66 & \cellcolor{red!25}0.15 
& \cellcolor{yellow!25}17.22 & \cellcolor{orange!25}0.57 & \cellcolor{red!25}0.24 
& \cellcolor{red!25}19.05 & \cellcolor{red!25}0.76 & \cellcolor{red!25}0.15 & \cellcolor{orange!25}{18.812} & \cellcolor{orange!25}0.675 & \cellcolor{red!25}0.160 \\

\bottomrule
\end{tabular}
 }
\vspace{-3mm}
\caption{Comparison of Cloth-HUGS with prior work on the NeuMan dataset \cite{jiang2022neuman} over \textit{human-only cropped regions}, using PSNR↑, SSIM↑, and LPIPS↓. Higher is better for PSNR and SSIM. Lower is better for LPIPS.}
\label{tab:neuman_human_full}
\end{table*}

\begin{table*}[t]
    \centering
    \small
    \adjustbox{max width=\textwidth}{
    \begin{tabular}{l | ccc | ccc | ccc | ccc | ccc | ccc || ccc}
        \toprule
        & \multicolumn{3}{c|}{377}
        & \multicolumn{3}{c|}{386}
        & \multicolumn{3}{c|}{387}
        & \multicolumn{3}{c|}{392}
        & \multicolumn{3}{c|}{393}
        & \multicolumn{3}{c||}{394}
        & \multicolumn{3}{c}{Average} \\
        \cmidrule(lr){2-4} \cmidrule(lr){5-7} \cmidrule(lr){8-10}
        \cmidrule(lr){11-13} \cmidrule(lr){14-16} \cmidrule(lr){17-19} \cmidrule(lr){20-22}

        Method 
            & PSNR & SSIM & LPIPS
            & PSNR & SSIM & LPIPS
            & PSNR & SSIM & LPIPS
            & PSNR & SSIM & LPIPS
            & PSNR & SSIM & LPIPS
            & PSNR & SSIM & LPIPS
            & PSNR & SSIM & LPIPS \\
        \midrule

        NeuralBody
            & 29.11 & \cellcolor{yellow!25}0.97 & \cellcolor{orange!25}0.04
            & 30.54 & \cellcolor{orange!25}0.97 & \cellcolor{yellow!25}0.05
            & 27.00 & \cellcolor{yellow!25}0.95 & 0.06
            & 30.10 & 0.96 & \cellcolor{orange!25}0.05
            & 28.61 & \cellcolor{orange!25}0.96 & \cellcolor{yellow!25}0.06
            & 29.10 & \cellcolor{orange!25}0.96 & 0.05
            & 29.076 & 0.962 & 0.052 \\

        HumanNeRF
            & 30.41 & \cellcolor{yellow!25}0.97 & \cellcolor{red!25}0.02
            & \cellcolor{yellow!25}33.20 & \cellcolor{red!25}0.98 & \cellcolor{orange!25}0.03
            & \cellcolor{yellow!25}28.18 & \cellcolor{orange!25}0.96 & \cellcolor{yellow!25}0.04
            & 31.04 & 0.97 & \cellcolor{red!25}0.03
            & 28.31 & \cellcolor{orange!25}0.96 & \cellcolor{orange!25}0.04
            & 30.31 & \cellcolor{orange!25}0.96 & \cellcolor{yellow!25}0.03
            & \cellcolor{yellow!25}30.241 & \cellcolor{yellow!25}0.967 & \cellcolor{yellow!25}0.032 \\

        MonoHuman
            & \cellcolor{yellow!25}30.77 & \cellcolor{red!25}0.98 & \cellcolor{red!25}0.02
            & 32.97 & \cellcolor{orange!25}0.97 & \cellcolor{orange!25}0.03
            & 27.93 & \cellcolor{orange!25}0.96 & \cellcolor{orange!25}0.03
            & \cellcolor{yellow!25}31.24 & \cellcolor{orange!25}0.97 & \cellcolor{red!25}0.03
            & \cellcolor{yellow!25}28.46 & \cellcolor{orange!25}0.96 & \cellcolor{red!25}0.03
            & \cellcolor{yellow!25}28.94 & \cellcolor{orange!25}0.96 & \cellcolor{orange!25}0.04
            & 30.052 & \cellcolor{yellow!25}0.967 & \cellcolor{orange!25}0.030 \\

        HUGS
            & \cellcolor{orange!25}30.80 & \cellcolor{red!25}0.98 & \cellcolor{red!25}0.02
            & \cellcolor{red!25}34.11 & \cellcolor{red!25}0.98 & \cellcolor{red!25}0.02
            & \cellcolor{red!25}29.29 & \cellcolor{red!25}0.97 & \cellcolor{orange!25}0.03
            & \cellcolor{orange!25}31.36 & \cellcolor{orange!25}0.97 & \cellcolor{red!25}0.03
            & \cellcolor{red!25}29.80 & \cellcolor{red!25}0.97 & \cellcolor{red!25}0.03
            & \cellcolor{orange!25}30.54 & \cellcolor{red!25}0.97 & \cellcolor{red!25}0.03
            & \cellcolor{orange!25}30.983 & \cellcolor{orange!25}0.973 & \cellcolor{red!25}0.028 \\

        \midrule
        \textbf{Ours}
            & \cellcolor{red!25}30.87 & \cellcolor{red!25}0.98 & \cellcolor{red!25}0.02
            & \cellcolor{orange!25}33.79 & \cellcolor{red!25}0.98 & \cellcolor{red!25}0.02
            & \cellcolor{orange!25}29.23 & \cellcolor{red!25}0.97 & \cellcolor{red!25}0.02
            & \cellcolor{red!25}31.74 & \cellcolor{red!25}0.98 & \cellcolor{red!25}0.03
            & \cellcolor{orange!25}29.64 & \cellcolor{red!25}0.97 & \cellcolor{red!25}0.03
            & \cellcolor{red!25}30.72 & \cellcolor{red!25}0.97 & \cellcolor{red!25}0.03
            & \cellcolor{red!25}30.998 & \cellcolor{red!25}0.975 & \cellcolor{red!25}0.028 \\

        \bottomrule
    \end{tabular}
    }
    \vspace{-3mm}
    \caption{Comparison of Cloth-HUGS (ours) with previous work on the ZJU-MoCap dataset \cite{peng2021neural}. Performance is evaluated using PSNR↑, SSIM↑, and LPIPS↓. Higher is better for PSNR and SSIM. Lower is better for LPIPS.}
    \vspace{-1mm}
    \label{tab:zjumocap}
\end{table*}

\begin{table*}[t]
    \centering
    \small
    \adjustbox{max width=\textwidth}{%
    \begin{tabular}{l | ccc | ccc | ccc | ccc | ccc | ccc || ccc}
        \toprule
        & \multicolumn{3}{c|}{Seattle}
        & \multicolumn{3}{c|}{Citron}
        & \multicolumn{3}{c|}{Parkinglot}
        & \multicolumn{3}{c|}{Bike}
        & \multicolumn{3}{c|}{Jogging}
        & \multicolumn{3}{c||}{Lab} 
        & \multicolumn{3}{c}{Average} \\
        \cmidrule(lr){2-4} \cmidrule(lr){5-7} \cmidrule(lr){8-10}
        \cmidrule(lr){11-13} \cmidrule(lr){14-16} \cmidrule(lr){17-19} \cmidrule(lr){20-22}

        Method
            & PSNR & SSIM & LPIPS
            & PSNR & SSIM & LPIPS
            & PSNR & SSIM & LPIPS
            & PSNR & SSIM & LPIPS
            & PSNR & SSIM & LPIPS
            & PSNR & SSIM & LPIPS
            & PSNR & SSIM & LPIPS \\
        \midrule

        {\footnotesize w/o Phys. losses}
            & 18.24 & \cellcolor{yellow!25}0.62 & \cellcolor{yellow!25}0.16
            & 18.36 & \cellcolor{orange!25}0.68 & \cellcolor{yellow!25}0.15
            & 18.95 & \cellcolor{yellow!25}0.69 & \cellcolor{red!25}0.15
            & \cellcolor{red!25}19.50 & \cellcolor{yellow!25}0.64 & \cellcolor{yellow!25}0.17
            & \cellcolor{orange!25}17.23 & \cellcolor{red!25}0.57 & \cellcolor{orange!25}0.25
            & 17.87 & \cellcolor{yellow!25}0.69 & \cellcolor{orange!25}0.21
            & 18.358 & 0.648 & 0.182\\

        {\footnotesize w/o Cloth LBS Reg.}
            & \cellcolor{orange!25}18.52 & \cellcolor{orange!25}0.63 & \cellcolor{red!25}0.14
            & 17.61 & 0.66 & 0.17
            & \cellcolor{yellow!25}19.06 & 0.66 & \cellcolor{orange!25}0.19
            & 19.20 & \cellcolor{yellow!25}0.64 & \cellcolor{orange!25}0.16
            & 17.05 & \cellcolor{red!25}0.57 & \cellcolor{red!25}0.24
            & 17.78 & \cellcolor{yellow!25}0.69 & \cellcolor{orange!25}0.21
            & 18.203 & 0.642 & 0.185 \\

        {\footnotesize ARAP loss only}
            & 18.35 & \cellcolor{yellow!25}0.62 & \cellcolor{orange!25}0.15
            & \cellcolor{orange!25}18.77 & \cellcolor{yellow!25}0.67 & \cellcolor{yellow!25}0.15
            & 19.05 & \cellcolor{orange!25}0.70 & \cellcolor{red!25}0.15
            & \cellcolor{yellow!25}19.35 & \cellcolor{orange!25}0.65 & \cellcolor{orange!25}0.16
            & \cellcolor{red!25}17.28 & \cellcolor{orange!25}0.56 & \cellcolor{orange!25}0.25
            & \cellcolor{yellow!25}18.84 & \cellcolor{orange!25}0.75 & \cellcolor{red!25}0.15
            & \cellcolor{orange!25}18.607 & 0.658 & \cellcolor{yellow!25}0.168 \\

        {\footnotesize Simulation loss only}
            & 17.95 & \cellcolor{yellow!25}0.62 & \cellcolor{yellow!25}0.16
            & 18.25 & \cellcolor{yellow!25}0.67 & \cellcolor{yellow!25}0.15
            & 18.98 & \cellcolor{orange!25}0.70 & \cellcolor{red!25}0.15
            & \cellcolor{orange!25}19.48 & \cellcolor{orange!25}0.65 & \cellcolor{orange!25}0.16
            & 17.18 & \cellcolor{red!25}0.57 & \cellcolor{red!25}0.24
            & 18.78 & \cellcolor{orange!25}0.75 & \cellcolor{red!25}0.15
            & 18.437 & \cellcolor{yellow!25}0.660 &\cellcolor{yellow!25}0.168 \\

        {\footnotesize Mask loss only}
            & \cellcolor{yellow!25}18.47 & \cellcolor{orange!25}0.63 & \cellcolor{red!25}0.14
            & \cellcolor{yellow!25}18.59 & \cellcolor{orange!25}0.68 & \cellcolor{orange!25}0.14
            & \cellcolor{orange!25}19.16 & \cellcolor{orange!25}0.70 & \cellcolor{red!25}0.15
            & 19.28 & \cellcolor{orange!25}0.65 & \cellcolor{orange!25}0.16
            & 17.13 & \cellcolor{red!25}0.57 & \cellcolor{orange!25}0.25
            & \cellcolor{orange!25}18.99 & \cellcolor{orange!25}0.75 & \cellcolor{red!25}0.15
            & \cellcolor{yellow!25}18.603 & \cellcolor{orange!25}0.663 & \cellcolor{orange!25}0.165 \\

        \midrule
        \textbf{Ours}
            & \cellcolor{red!25}18.68 & \cellcolor{red!25}0.65 & \cellcolor{red!25}0.14
            & \cellcolor{red!25}19.12 & \cellcolor{red!25}0.70 & \cellcolor{red!25}0.13
            & \cellcolor{red!25}19.32 & \cellcolor{red!25}0.71 & \cellcolor{red!25}0.15
            & \cellcolor{orange!25}19.48 & \cellcolor{red!25}0.66 & \cellcolor{red!25}0.15
            & \cellcolor{yellow!25}17.22 & \cellcolor{red!25}0.57 & \cellcolor{red!25}0.24
            & \cellcolor{red!25}19.05 & \cellcolor{red!25}0.76 & \cellcolor{red!25}0.15
            & \cellcolor{red!25}18.812 & \cellcolor{red!25}0.675 & \cellcolor{red!25}0.160 \\

        \bottomrule
    \end{tabular}
    }
    \vspace{-3mm}
    \caption{\textbf{Ablation study.} The performance is evaluated over human-only bounding box regions in NeuMan dataset \cite{jiang2022neuman} using PSNR↑, SSIM↑, and LPIPS↓. Higher is better for PSNR and SSIM. Lower is better for LPIPS.}
    \label{tab:ablation}
    \vspace{-2mm}
\end{table*}


\begin{table}[t]
\centering
\resizebox{\columnwidth}{!}{%
\begin{tabular}{l|c|c|c|c|c|c||c}
\toprule
\multicolumn{8}{c}{\textbf{Full Scene}} \\
\midrule
Method 
& Seattle & Citron & Parking & Bike & Jogging & Lab & Average \\
\midrule
NeuMan 
& \normalsize\cellcolor{yellow!25}70.61 
& \normalsize\cellcolor{yellow!25}93.83
& \normalsize\cellcolor{yellow!25}96.20 
& \normalsize\cellcolor{yellow!25}47.97 
& \normalsize\cellcolor{orange!25}98.08 
& \normalsize\cellcolor{yellow!25}87.66 & \cellcolor{yellow!25}82.392 \\
HUGS 
& \normalsize\cellcolor{orange!25}59.69 
& \normalsize\cellcolor{orange!25}66.76 
& \normalsize\cellcolor{orange!25}93.53 
& \normalsize\cellcolor{orange!25}45.48 
& \normalsize\cellcolor{red!25}97.96 
& \normalsize\cellcolor{orange!25}48.74 & \cellcolor{orange!25}68.693\\
Ours 
& \normalsize\cellcolor{red!25}50.51 
& \normalsize\cellcolor{red!25}65.90 
& \normalsize\cellcolor{red!25}85.42 
& \normalsize\cellcolor{red!25}39.08 
& \normalsize\cellcolor{yellow!25}99.66 
& \normalsize\cellcolor{red!25}47.97 & \cellcolor{red!25}64.757 \\
\midrule
\midrule
\multicolumn{8}{c}{\textbf{Human-only regions}} \\
\midrule
Method 
& Seattle & Citron & Parking & Bike & Jogging & Lab & Average \\
\midrule
NeuMan 
& \normalsize\cellcolor{yellow!25}129.61 
& \normalsize\cellcolor{yellow!25}100.54 
& \normalsize\cellcolor{yellow!25}138.34 
& \normalsize\cellcolor{yellow!25}94.01 
& \normalsize\cellcolor{yellow!25}116.71 
& \normalsize\cellcolor{yellow!25}152.24 & \cellcolor{yellow!25}121.908 \\
HUGS 
& \normalsize\cellcolor{orange!25}103.20 
& \normalsize\cellcolor{orange!25}99.38 
& \normalsize\cellcolor{red!25}96.72 
& \normalsize\cellcolor{orange!25}66.38 
& \normalsize\cellcolor{orange!25}107.41 
& \normalsize\cellcolor{orange!25}105.43 & \cellcolor{orange!25}96.420 \\
Ours 
& \normalsize\cellcolor{red!25}93.10 
& \normalsize\cellcolor{red!25}91.15 
& \normalsize\cellcolor{orange!25}100.02 
& \normalsize\cellcolor{red!25}60.87 
& \normalsize\cellcolor{red!25}92.53 
& \normalsize\cellcolor{red!25}104.80 & \cellcolor{red!25}90.412 \\
\bottomrule
\end{tabular}
}
\vspace{-3mm}
\caption{FID↓ comparison on full scene (top) and human-only regions (bottom) across six NeuMan \cite{jiang2022neuman} sequences.}
\label{tab:fid_comparison}
\end{table}

\vspace{-3mm}
\section{Results}
\label{sec:results}
\vspace{-2mm}
In this section, we provide qualitative and quantitative comparisons, followed by the ablation study results.

\vspace{-2mm}
\subsection{Qualitative Results}
\label{sec:qualitative_results}
\vspace{-1mm}

Fig.~\ref{fig:qualitative} compares Cloth-HUGS with HUGS~\cite{kocabas2023hugs} and NeuMan~\cite{jiang2022neuman} on unseen NeuMan test sequences.
Across subjects, poses, and apparel, Cloth-HUGS produces sharper facial details, cleaner body-cloth boundaries, and more accurate cloth geometry.

In the first row, Cloth-HUGS preserves finer wrinkles near the T-shirt hem,
whereas HUGS appears smoother and NeuMan shows pose-related artifacts. 
In the second row, Cloth-HUGS resolves hand-shirt separation and occlusions, while HUGS blends regions, and NeuMan misplaces the arm. Cloth-HUGS also preserves clearer pant-level wrinkle cues.
Overall, Cloth-HUGS achieves the strongest perceptual fidelity and geometric coherence, enabled by explicit body-cloth separation and physics-guided regularization.

\vspace{-2mm}
\subsection{Quantitative Results}
\label{sec:quantitative_results}
\vspace{-1mm}

We compare Cloth-HUGS with NeRF-T~\cite{li2021neural}, HyperNeRF~\cite{park2021hypernerf}, Vid2Avatar~\cite{guo2023vid2avatar}, NeuMan~\cite{jiang2022neuman}, and HUGS~\cite{kocabas2023hugs}. For human-only evaluation, NeRF-T and HyperNeRF are omitted due to the lack of explicit human modeling.

As shown in Table~\ref{tab:neuman_full}, NeRF-T and HyperNeRF perform poorly on articulated humans, while NeuMan and Vid2Avatar improve results but miss fine-scale clothing detail. HUGS treats cloth as part of the body, often blurring boundaries under motion. By explicitly separating body and cloth, Cloth-HUGS produces sharper folds and improved temporal stability. Quantitatively, Cloth-HUGS achieves higher PSNR, comparable SSIM, and substantially lower LPIPS (0.157 $\rightarrow$ 0.113, 28\%) on full scenes. On human-only crops (Table~\ref{tab:neuman_human_full}), LPIPS further improves from 0.185 to 0.160 while maintaining competitive PSNR/SSIM. FID is also reduced (Table~\ref{tab:fid_comparison}), from 68.693 to 64.757 (5.73\%) for full scenes and from 96.420 to 90.412 (6.23\%) for human-only evaluation.

The results of ZJU-MoCap~\cite{peng2021neural} are reported in Table~\ref{tab:zjumocap}. Cloth-HUGS achieves the best overall PSNR, SSIM, and LPIPS, with larger gains under non-rigid motion and complex apparel. Body–cloth disentanglement, physics-inspired regularization, and depth-aware compositing preserve sharp details and stable deformation, demonstrating improved perceptual realism over prior Gaussian-based methods.

\begin{figure}[]
    \centering
    \includegraphics[width=\columnwidth]{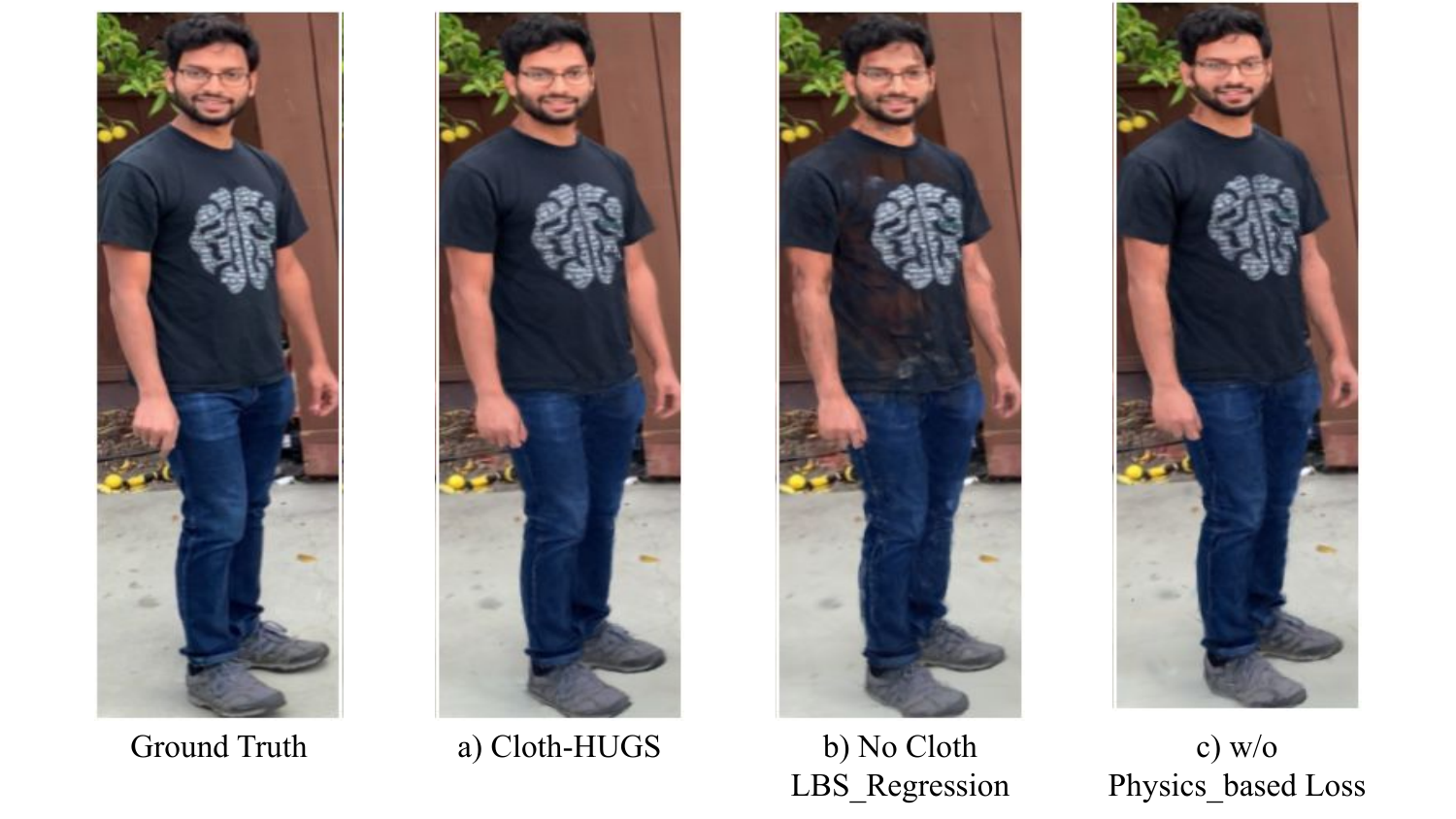}
    \vspace{-8mm}
    \caption{
    \textbf{Ablation study.} 
Comparison of key components in Cloth-HUGS: 
(a) full model, 
(b) without cloth LBS regularization, 
and (c) without physics-based losses. 
}
    \label{fig:ablation}
    \vspace{-4mm}
\end{figure}

\vspace{-2mm}
\subsection{Ablation Studies}
\vspace{-1mm}

We evaluate the contributions of the individual components of our model in Table~\ref{tab:ablation} and visualize representative
results in Fig.~\ref{fig:ablation}. 
Results demonstrate that the removal of physics-inspired losses degrades PSNR and SSIM while increasing LPIPS, indicating reduced perceptual quality. Ablating cloth LBS regularization causes unstable deformations and poor body attachment. The full model achieves optimal performance (PSNR: 18.812, SSIM: 0.675, LPIPS: 0.160), confirming that physics-based supervision, LBS regularization, and depth-aware compositing are complementary and essential for photorealistic clothed human reconstruction.
\vspace{-2mm}
\section{Conclusion}
\label{sec:conclusion}
\vspace{-1mm}
We introduced a neural rendering framework that explicitly models cloth as independent geometric entities while maintaining skeletal coupling through LBS weight regularization. A depth-aware multi-pass renderer ensures accurate occlusion among body, cloth, and scene layers. Through ablation studies, we find that well-regularized LBS weights provide a strong geometric prior for stable cloth attachment and motion consistency, forming a robust foundation for clothed human rendering. By integrating cloth priors from SNUG~\cite{santesteban2022snug} with neural rendering, our method enables plausible cloth deformation that follows body motion while retaining the efficiency of 3D Gaussian Splatting~\cite{Kerbl20233DGaussianSplatting}. Experiments on NeuMan~\cite{jiang2022neuman} and ZJU-MoCap~\cite{peng2021neural} datasets validate that well-regularized LBS weights provide strong geometric guidance for realistic clothed human rendering.

\bibliographystyle{IEEEbib}
\bibliography{strings,refs}

\end{document}